# MSHCNet: Multi-Stream Hybridized Convolutional Networks with Mixed Statistics in Euclidean/Non-Euclidean Spaces and Its Application to Hyperspectral Image Classification

Shuang He, Haitong Tang, Xia Lu, Hongjie Yan, Nizhuan Wang*

*Abstract*—It is well known that hyperspectral images (HSI) contain rich spatial-spectral contextual information, and how to effectively combine both spectral and spatial information using DNN for HSI classification has become a new research hotspot. Compared with CNN with square kernels, GCN have exhibited exciting potential to model spatial contextual structure and conduct flexible convolution on arbitrarily irregular image regions. However, current GCN only using first-order spectral-spatial signatures can result in boundary blurring and isolated misclassification. To address these, we first designed the graph-based second-order pooling (GSOP) operation to obtain contextual nodes information in non-Euclidean space for GCN. Further, we proposed a novel multi-stream hybridized convolutional network (MSHCNet) with combination of first and second order statistics in Euclidean/non-Euclidean spaces to learn and fuse multi-view complementary information to segment HSIs. Specifically, our MSHCNet adopted four parallel streams, which contained G-stream, utilizing the irregular correlation between adjacent land covers in terms of first-order graph in non-Euclidean space; C-stream, adopting convolution operator to learn regular spatial-spectral features in Euclidean space; N-stream, combining first and second order features to learn representative and discriminative regular spatial-spectral features of Euclidean space; S-stream, using GSOP to capture boundary correlations and obtain graph representations from all nodes in graphs of non-Euclidean space. Besides, these feature representations learned from four different streams were fused to integrate the multi-view complementary information for HSI classification. Finally, we evaluated our proposed MSHCNet on three hyperspectral datasets, and experimental results demonstrated that our method significantly outperformed state-of-the-art eight methods.

This work was supported by National Natural Science Foundation of China (No. 41506106, 61701318, 82001160), Project of "Six Talent Peaks" of Jiangsu Province (No. SWYY-017), and Project of Huaguoshan Mountain Talent Plan - Doctors for Innovation and Entrepreneurship, Jiangsu University Superior Discipline Construction Project Funding Project (PAPD).

Shuang He, Haitong Tang, Xia Lu, Nizhuan Wang are with School of Marine Technology and Geomatics, School of Computer Engineering, Jiangsu Key Laboratory of Marine Bioresources and Environment, and Jiangsu Key Laboratory of Marine Biotechnology, Jiangsu Ocean University, People's Republic of China (e-mails: kyrohe95@gmail.com, httang1224@gmail.com, lux2008000070@jou.edu.cn, wangnizhuan1120@gmail.com ).

Hongjie Yan is with Department of Neurology, Affiliated Lianyungang Hospital of Xuzhou Medical University, People's Republic of China (e-mail: yanhjns@gmail.com).

* denotes the corresponding author.

## I. INTRODUCTION

Hyperspectral remote sensing is a prominent multidimensional information collection technology that successfully combines imaging and spectral techniques [1, 2]. It is widely utilized in the areas of fine farming [3], environmental monitoring [4], soil prediction [5], etc. Hyperspectral image (HSI) classification [6, 7] is a technique by automatically identifying characteristics to assign a single category label to each pixel.

To improve HSI classification performance, diverse kinds of approaches have been proposed over the past decades, including traditional machine leaning-based (ML-based) and deep learning-based (DL-based) methods [8-12]. Generally speaking, traditional ML-based methods like support vector machine (SVM) [13-14], random forest (RF) [15], sparse representation-based classifier (SRC) [16], and K-means [17] are popular for classification of HSIs. However, the aforementioned methods are all based on the handcraft spectral-spatial features, and have difficulties such as too little labeling information available and insufficient extraction of essential features.

To address the aforementioned issues in traditional ML-based methods, recently, deep learning [18, 19] is extensively applied to HSI classification and its great representational capacity has gained increasing attention. The main reason is that deep learning can automatically extract features from the lower to upper level, which solves the problem of insufficient feature extraction by traditional ML-based approaches to a certain extent. Typical DL-based frameworks for HSI classification include stacked autoencoder (SAE) [20], recurrent neural network (RNN) [21, 22], convolutional neural network (CNN) [12, 23, 25-34], and graph convolutional network (GCN) [24, 35-39]. Specifically, CNN recently has become the dominant architecture for HSI classification, which can be categorized as 1D CNN [23], extracting the spectral features along the radiometric dimension, 2D CNN [26, 27], extracting features directly from the inputs and modeling the spatial information with 2D convolutions, 3D CNN [28-31], using 3D convolution kernel for learning spectral-spatial feature, and CNN-based ones with high-order statistics [33, 34], aiming to aggregate and exploit the correlation information. GCNs [38, 39], as emerging network structures, are able to escape the limitations of grid samples and generally adapt to different local regions, especially on class boundaries.

TABLE I
REPRESENTATIVE MODELS WITH THEIR ADVANTAGE AND DISADVANTAGE FOR HSI CLASSIFICATION,
WHERE ✓ REPRESENTS ADVANTAGE AND ✗ DENOTES DISADVANTAGE.

| Methods | | Representatives | Accuracy | Memory Consumption | Time Cost | Representability | Generalization |
|---|---|---|---|---|---|---|---|
| Traditional ML-based | SVM | Bovolo *et al.* [13] | ✓ | ✗ | ✗ | ✗ | ✓ |
| | RF | Xia *et al.* [15] | ✗ | ✓ | ✗ | ✗ | ✗ |
| | SRC | He *et al.* [16] | ✗ | ✗ | ✗ | ✓ | ✗ |
| | K-Means | Ling *et al.* [17] | ✗ | ✓ | ✓ | ✗ | ✗ |
| DL-based | SAE | Chen *et al.* [20] | ✗ | ✗ | ✗ | ✗ | ✓ |
| | RNN | Mou *et al.* [21] | ✗ | ✓ | ✗ | ✓ | ✗ |
| | CNN-based 1D CNN | Hu *et al.* [23] | ✗ | ✓ | ✗ | ✗ | ✗ |
| | CNN-based 2D CNN | Makantasis *et al.* [26] | ✓ | ✓ | ✗ | ✗ | ✗ |
| | CNN-based 3D CNN | Chen *et al.* [28] | ✓ | ✗ | ✗ | ✓ | ✗ |
| | CNN-based high-order | Xue *et al.* [34] | ✓ | ✗ | ✗ | ✓ | ✓ |
| | GCN-based | Shahraki *et al.* [35] | ✓ | ✗ | ✓ | ✓ | ✓ |

Table I summarized the performance of some representative models for HSI classification in terms of their accuracy, memory consumption, time cost, representability, and generalization. From Table I, it showed that traditional ML-based methods, SAE and RNN lacked of enough representability, generalization, or accuracy. The main reasons for that were that the features were under-extracted and easily over-fitting. Although these CNN-based methods had achieved satisfying performance in feature learning and classification, a large amount of labeled samples were required to train these architectures under a supervised classification task. Due to the difficulty in obtaining labeled training samples, they had limited generalization and high time cost. While high-order statistics could address the generalization problems of CNN-based methods, it could only operate on regular regions with fixed size and weights. As we can see, from Table I, that GCN-based methods had strong accuracy, short time cost, strong representability and well generalization. Nevertheless, there were some potential defects of the existing GCNs for HSI classification. Firstly, since hyperspectral data were often contaminated by salt and pepper noise, the initial input graph may not be accurate, which cannot represent their intrinsic similarities. Furthermore, traditional GCNs only utilized the first-order spectral signatures, without considering the second-order information that was salient local information embedded in the original HSI, potentially resulting in isolated false predictions on the HSI. Last but not least, the spatial memory computational complexity of the existing GCN was a significant bottleneck in HSI classification task, which would be unacceptable when the number of pixels got too large.

To resolve these issues, inspired by our previous work [40], in this paper we propose a novel multi-stream hybridized convolutional network (MSHCNet) with mixed first and second statistics in both Euclidean and non-Euclidean spaces to learn and fuse multi-view information for end-to-end HSI classification, which starts with four parallel branches, namely C-stream, G-stream, N-stream, and S-stream, to learn independently multi-scale feature representations, respectively. That is, C-stream adopts convolutional neural network to learn regular spectral-spatial information from HSI; the N-stream involves a second-order pooling (SOP) operator to model the discriminative and representative spectral-spatial local features; the G-stream adopts graph convolutional network to capture the irregular correlation between adjacent land covers and model distinctive spatial structural details; the S-stream adopts a graph-based second-order pooling (GSOP) operator to capture spatial boundary correlations and obtain graph representations from all nodes, which can further describe local features and boundary information correlations. Then, these distinctive representations of HSI from four parallel streams are further fused by concatenation strategy to learn complementary multi-view information for HSI classification.

More specifically, the main contributions of this work can be summarized as follows:

1) We propose a novel multi-stream hybridized convolutional network (i.e., MSHCNet) by integrating four parallel streams for discriminative features fusion to label each pixel in HSI. The G-stream is presented to learn the HSI spatial contextual structure on irregular regions; The C-stream is applied to extract Euclidean spatial-spectral feature on regular areas; Furthermore, the N-stream and S-stream ensure that our model achieve local and boundary features in Euclidean and non-Euclidean spaces.
2) The mixed statistics with first/second order features are presented to aggregate and propagate boundary and contextual information by hybridized order statistics, where the first order statistics can capture contextual spectral-spatial information, and second order statistics can improve the aggregation and representation ability to generate more representative and discriminative features.
3) Our MSHCNet is evaluated on three typical hyperspectral image datasets, and the experimental results show that our MSHCNet significantly outperforms the other eight state-of-the-art HSI classification methods.

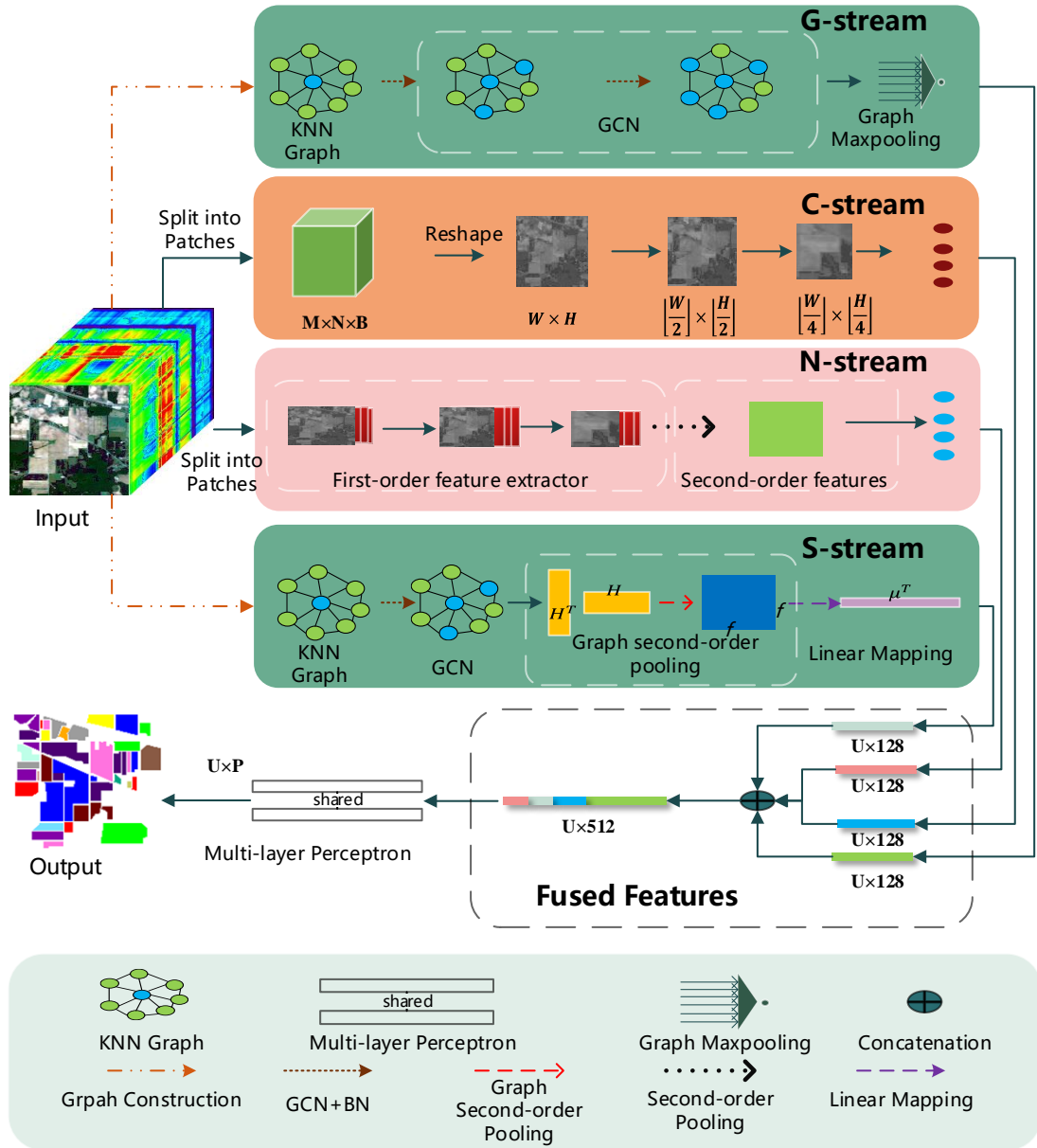

Fig. 1. Structure of the proposed MSHCNet. The network takes raw hyperspectral data as inputs, and adopts four independent convolutional and graph convolutional streams (i.e., G-stream, C-stream, N-stream and S-stream) to learn discriminative and complementary spatial-spectral representations from different features. Then, the high level feature produced by each stream are fused for final node classification.

The remainder of this article is organized as follows. Section II details the proposed MSHCNet. Extensive experiments and the corresponding analysis are given in Section III. Finally, Section IV and V conclude the proposed method and discuss the related future works.

## II. THE PROPOSED METHOD

### A. Brief Introduction of GCN

In 2005, Gori *et al.* [41] first introduced the notion of graph neural network (GNN), which had the benefit over CNN in that it can work on graph-structured non-Euclidean data. Subsequently, Scarselli *et al.* [42] made GNN trainable by a supervised learning algorithm for practical data. Bruna *et al.* [43] proposed the first GCN based on spectral property, which convolved on the neighborhood of every graph node and produced a node-level output. After that, many extensions of graph convolution had been investigated and achieved advanced results [44, 45]. Based on their work, Kipf and Welling [46] proposed a fast approximation localized convolution, which made the GCN able to encode both graph structure and node features.

**Algorithm 1** The Proposed G-stream in MSHCNet
---
**Input:** Input original image $\mathbf{I}_m$; ground truth $\mathbf{I}_{gt}$; the number of nearest neighbors $K$;
1  Use KNN-G to construct graph G;
2  Calculate the Laplacian metrics $\mathbf{L}_{sym}$ according to $\mathbf{L}_{sym} = \mathbf{I} - \mathbf{D}^{-\frac{1}{2}}\mathbf{A}\mathbf{D}^{-\frac{1}{2}}$;
3  **for** $l = 1$ to $N$ **do**
4    Use Batch Normalization to normalize input;
5    Calculate the outputs of the $l$th layer $\boldsymbol{h}^{(l)}$ according to Eq.1;
6    Update the GCN layer output $\boldsymbol{h}^{(l+1)}$;
7  **end do**
8  Use max-pooling on graph to calculate the features $\mathbf{f}$ according Eq.2;
9  Calculate the network output feature maps $\boldsymbol{H}_G = \text{reshape}(\mathbf{f})$;
**Output:** Network output $\boldsymbol{H}_G$.

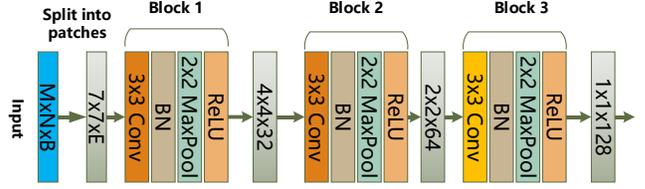

Fig. 2. Detailed network configuration in each layer of C-stream in MSHCNet. The proposed C-stream is generated by sequentially stacking three blocks.

In their work, GCN was simplified by a first-order approximation of graph spectral convolution, which contributed to more efficient filtering operations. Moreover, the neighborhood size in their method was also fixed and thus the spectral-spatial information in different local regions cannot be flexibly captured. To cope with the aforementioned issues, we proposed a novel multi-stream hybridized convolutional network (MSHCNet) which extracted the spectral-spatial information from different views and fused multi-view spectral-spatial information in both Euclidean and non-Euclidean spaces for HSI classification.

*B. Multi-Stream Architecture of MSHCNet*

Given an HSI classification problem, we define the input of our MSHCNet as matrix $\mathbf{I}_m$ with size of M×N×B. That is, each specific HSI is described by B bands. As illustrated in Fig. 1, our MSHCNet starts with a four-stream architecture, which adopts a G-stream, a C-stream, an N-stream, and an S-stream to learn the discriminative spatial-spectral representations from the hyperspectral data. Thereafter, the features produced by these four complementary streams are further fused to learn high-level representation for final classification. The output of MSHCNet is an [M×N, P] matrix, with each row denoting the probabilities of the respective pixel belonging to P different classes.

*1) G-stream*

Our G-stream is designed to learn the basic irregular feature with first order statistics in non-Euclidean space from HSI. As previously shown in Fig. 1, given the input of HSI, a series of graph convolutional layers are successively applied in the forward path to extract multi-scale features from each channel. In each layer of G-stream, a K-nearest neighbors graph (KNN-G) [47] $G$ is first constructed for the $M \times N$ pixels in terms of the input features. Specifically, for each pixel (i.e., a central node), we search its $K$ nearest pixels with the minimum Euclidean distance in spatial feature space. Let the resulting graph be $G = (\mathbf{V}, \mathbf{E})$, where $\mathbf{V} = \{m_1, m_2, \ldots, m_n\}$ and $\mathbf{E} \subseteq |\mathbf{V}| \times |\mathbf{V}|$ represent the set of nodes and the set of edges (defined by KNN connectivity), respectively. For each node $m_i \in \mathbf{V}$, we denote its K-nearest neighbors as $N(i)$.

Specifically, we adopt a graph convolutional layer to aggregate the calibrated neighborhood information to each center. The correspondingly spectral feature aggregation is followed by:

$$\boldsymbol{h}_i^{l+1} = \sum_{m_i \in \mathrm{N}(i)} \mathbf{L}_{sym} \times \widehat{\boldsymbol{h}}_i^l \times \mathbf{W}^l + \boldsymbol{b}_i^l, \quad (1)$$

where the $\boldsymbol{h}_i^{l+1}$ indicates the updated feature of center $m_i$, i.e., the input feature of the $(l + 1)$-th layer. In Eq. (1), the indicators $\mathbf{W}^l$, $\boldsymbol{b}_i^l$, $\widehat{\boldsymbol{h}}_i^l$ and $\mathbf{L}_{sym}$ represent the weights, biases, the updated nearest-neighbor representations, and the symmetric normalized Laplacian matrix, respectively. Thereafter, we apply the max-pooling on all neighbors' calibrated features to produce the boundary representation for the respective center, which can be formulated as

$$\mathbf{f} = maxpooling\{\boldsymbol{h}_i^{l+1}, \forall m_i \in N(i)\}, \quad (2)$$

where $\mathbf{f}$ represents the output features and $\boldsymbol{h}_i^{l+1}$ indicates the output of the GCN final layers, respectively. The detailed process is summarized in Algorithm 1.

*2) C-stream*

Although the G-stream can learn the basic topological relations and features from the HSI, it cannot sensitively distinguish between adjacent pixels belonging to different classes (e.g., boundaries of each crop). Therefore, as complementary to the G-stream for the accurate HSI classification, we further design a C-stream to extract distinctive spatial-spectral features of HSIs in Euclidean space.

Our C-stream consists of a series of CNN layers, max-pooling layers and batch normalization (BN) layers. Notably, we force each layer in the C-stream similar to 2D CNN, which contains three 2D convolutional blocks. Each convolutional block involves a 2D convolutional layer, a BN layer, a max-pooling layer, and a ReLU activation layer. Moreover, the receptive fields along the spatial and spectral domains for each convolutional layer are $3 \times 3 \times 32$, $3 \times 3 \times 64$, and $1 \times 1 \times 128$, respectively. The detailed structure is shown in Fig. 2.

It is known that different network architecture can be capable of extracting distinctive representations of HSIs. The C-stream can enhance the discrimination ability of spatial and spectral features in Euclidean space in contrast to G-stream.

*3) N-stream*

It is well known that the second-order pooling (SOP) can apply to generate the representative features in Euclidean space such as boundary compact features in both the spectral-spatial

---

**Algorithm 2** The Proposed N-stream in MSHCNet

**Input:** Input image and corresponding labels;
1. Calculate the first-order features $H_{first}$;
2. Batch normalize the $H_{first}$;
3. Calculate $H_{SOP}$ according to Eq.3;
4. Conduct vectorization to obtain $H_N$;

**Output:** Network output $H_N$.

---

domains. Therefore, our N-stream in MSHCNet consists of first-order feature extractor and SOP operator. To calculate the second-order statistical representations of the obtained first-order features, we define the SOP operator as follow according to [34],

$$H_{SOP} = H_{first}^T H_{first}, \quad (3)$$

where $H_{SOP}$ is a real symmetric matrix, and $H_{first}$ represents the first-order features by the first-order feature extractor [36].

It is obvious that N-stream could rely heavily on SOP operator to focus on the correlations and compact features between spectral over identical spatial locations. Based on the above description, the detailed process of the proposed N-

---

**Algorithm 3** The Proposed S-stream in MSHCNet

**Input:** Input original image $I_m$; the number of nearest neighbors $K$;
1. Calculate the $h^{(l+1)}$ according to Algorithm 1;
2. Calculate the graph second-order pooling features $H_{GSOP}$ according to Eq. 4;
3. Conduct vectorization to obtain $H_S$;

**Output:** Network output $H_S$.

---

stream is summarized in Algorithm 2.

*4) S-stream*

To extract fine-grained boundary correlation representations in local areas of HSI, we additionally design a S-stream summarized in Algorithm 3, which can sensitively capture the discriminative second-order statistics in the graphs.

Our S-stream mainly involves a second-order pooling for graph (GSOP) operator taking the form

$$H_{GSOP} = \sum_{i=1}^{n} h_i h_i^T = H_G^T H_G, \quad (4)$$

where $H_{GSOP}$ is a real symmetric matrix, viewed as an $f^2$-dimensional graph representation vector [48]. $i = 1,2,\cdots,n$ are indexes of $n$ vertex nodes of the graph. $H_G$ represents the node representations by GCN from Algorithm 1.

---

**Algorithm 4** The Proposed MSHCNet for HSI Classification

**Input:** Input HSI image and corresponding labels;
**Begin**
1. Calculate $H_G$ according to Algorithm 1;
2. Calculate the C-stream output $H_C$ according to Fig.2;
3. Calculate $H_N$ according to Algorithm 2;
4. Calculate $H_S$ according to Algorithm 3;
5. Concatenate feature maps $[H_G, H_C, H_N, H_S]$;
6. Conduct corresponding label prediction by Eq. 5;
7. Calculate the cross-entropy loss $L$ according to Eq. 6;
**End**
**Output:** Predicted labels for corresponding pixels.

---

Notice that the second-order pooling naturally fits the goal and requirements of graph pooling, and can be able to capture the powerful correlation among spatial-spectral features.

*5) Feature Fusion and Classification*

Assuming that the G-stream, C-stream, N-stream and S-stream have learned completely different feature representations from four complementary views, fusing their outputs can enable the overall network to comprehensively identify the detailed structure of HSIs. As shown in Fig. 1, the feature matrices from four complementary views are concatenated, where the multi-layer perceptron [49] (MLP) is used to generate the classification output matrix **O** with the dimension $[M \times N, P]$, with each row denoting the specific pixels belonging to $P$ different classes, which can be formulated as:

$$O = MLP(H_G \oplus H_C \oplus H_N \oplus H_S), \quad (5)$$

where $H_G$, $H_C$, $H_N$ and $H_S$ represent the corresponding view features, respectively, and $\oplus$ denotes the concatenation operator. Then, we train MSHCNet with cross-entropy classification loss, which can be formulated as:

$$L = -\sum_{i=1}^{M \times N} \sum_{p=1}^{P} O_{ip} \log y_{ip}, \quad (6)$$

where $O_{ip}$ and $y_{ip}$ denote the outputs and the ground-truth labeling probability for $p$-th class, respectively. Thus, based on the above description, the detailed process of the proposed MSHCNet is summarized in Algorithm 4.

TABLE II
LAND-COVER CLASSES OF THE INDIAN PINES DATASET,
WITH THE AMOUNTS OF TRAINING AND TEST DATA FOR EACH CLASS.

| Class No. | Class Name | Training | Test |
|---|---|---|---|
| 1 | Alfalfa | 15 | 39 |
| 2 | Corn-notill | 50 | 1384 |
| 3 | Corn-mintill | 50 | 784 |
| 4 | Corn | 50 | 184 |
| 5 | Grass-pasture | 50 | 447 |
| 6 | Grass-trees | 50 | 697 |
| 7 | Grass-pasture-moved | 15 | 11 |
| 8 | Hay-windrowed | 50 | 439 |
| 9 | Oats | 15 | 5 |
| 10 | Soybeans-notill | 50 | 918 |
| 11 | Sobeans-mintill | 50 | 2418 |
| 12 | Sobeans-clean | 50 | 564 |
| 13 | Wheat | 50 | 162 |
| 14 | Woods | 50 | 1244 |
| 15 | Bldg-grass-tree-driver | 50 | 330 |
| 16 | Stone-steel-towers | 50 | 45 |
| | TOTAL | 695 | 9671 |

TABLE III
LAND-COVER CLASSES OF THE PAVIA UNIVERSITY DATASET,
WITH THE AMOUNTS OF TRAINING AND TEST DATA FOR EACH CLASS.

| Class No. | Class Name | Training | Test |
|---|---|---|---|
| 1 | Asphalt | 548 | 6631 |
| 2 | Meadows | 540 | 18649 |
| 3 | Gravel | 392 | 2099 |
| 4 | Tress | 524 | 3064 |
| 5 | Metal-sheets | 265 | 1345 |
| 6 | Bare-soil | 532 | 5029 |
| 7 | Bitumen | 375 | 1330 |
| 8 | Bricks | 514 | 3682 |
| 9 | Shadows | 231 | 947 |
| | TOTAL | 3921 | 42776 |

TABLE IV
LAND-COVER CLASSES OF THE HOUSTON2013 DATASET,
WITH THE AMOUNTS OF TRAINING AND TEST DATA FOR EACH CLASS.

| Class No. | Class Name | Training | Test |
|---|---|---|---|
| 1 | Healthy grass | 198 | 1053 |
| 2 | Stressed grass | 190 | 1064 |
| 3 | Synthetic grass | 192 | 505 |
| 4 | Trees | 188 | 1056 |
| 5 | Soil | 186 | 1056 |
| 6 | Water | 182 | 143 |
| 7 | Residential | 196 | 1072 |
| 8 | Commercial | 191 | 1053 |
| 9 | Road | 193 | 1059 |
| 10 | Highway | 191 | 1036 |
| 11 | Railway | 181 | 1054 |
| 12 | Parking Lot1 | 192 | 1041 |
| 13 | Parking Lot2 | 184 | 285 |
| 14 | Tennis court | 181 | 247 |
| 15 | Running track | 187 | 473 |
| | TOTAL | 2832 | 12197 |

## III. EXPERIMENTS AND ANALYSIS

### A. Hyperspectral Data Description

The performance of the proposed MSHCNet is evaluated on three public datasets, i.e., the Indian pine, the Pavia University and the university of Houston campus, which will be briefly introduced below.

*1) Indian Pines Hyperspectral Dataset*

The first hyperspectral data set was acquired by the Airborne Visible/Infrared Imaging Spectrometer (AVIRIS) sensor over northwestern Indiana, USA. The scene comprises of 145 × 145 pixels with a ground sampling distance (GSD) of 20 m and 220 spectral bands in the wavelength range from 400 to 2500 nm, at 10-nm spectral resolution. We retained 200 channels by removing 20 noisy and water absorption bands, i.e., 104-108, 150-163, and 220. Table II listed 16 main land-cover categories involved in this studied scene, as well as the number of training and testing samples used for the classification task. Correspondingly, Fig. 3 showed a false-color image of this scene and the spatial distribution of training and test samples.

*2) Pavia University Hyperspectral Dataset*

The second data set was acquired by the Reflective Optics Spectrographic Imaging System (ROSIS) sensor over the urban area of the University of Pavia, Italy. The number of wavelength bands in the acquired image is 103 ranged from 0.43 to 0.86 $\mu m$, and the image size is 610 × 340, with a very high spatial resolution of 1.3 m. The false-color composite of the University of Pavia image and the corresponding ground-truth map were shown in Fig. 3. The ground-truth map totally included nine classes as shown in Table III.

*3) Houston2013 Dataset*

This data set was used for the 2013 IEEE GRSS data fusion contest, and was collected using the ITRES CASI-1500 sensor over the campus of University of Houston and its surrounding rural areas in TX, USA. The image size is 349 × 1905 pixels with 144 spectral bands ranging from 364 to 1046 nm, at 10-nm spectral resolution. It should be noted that the used data set is a cloud-free hyperspectral product, processes by removing some small structures according to the illumination-related threshold map computed based on the spectral signatures. Table IV listed 15 challenging land-cover categories and the training and test sets. In Fig. 3, we showed a false-color image of the hyperspectral scene and the corresponding distribution of the training and test samples.

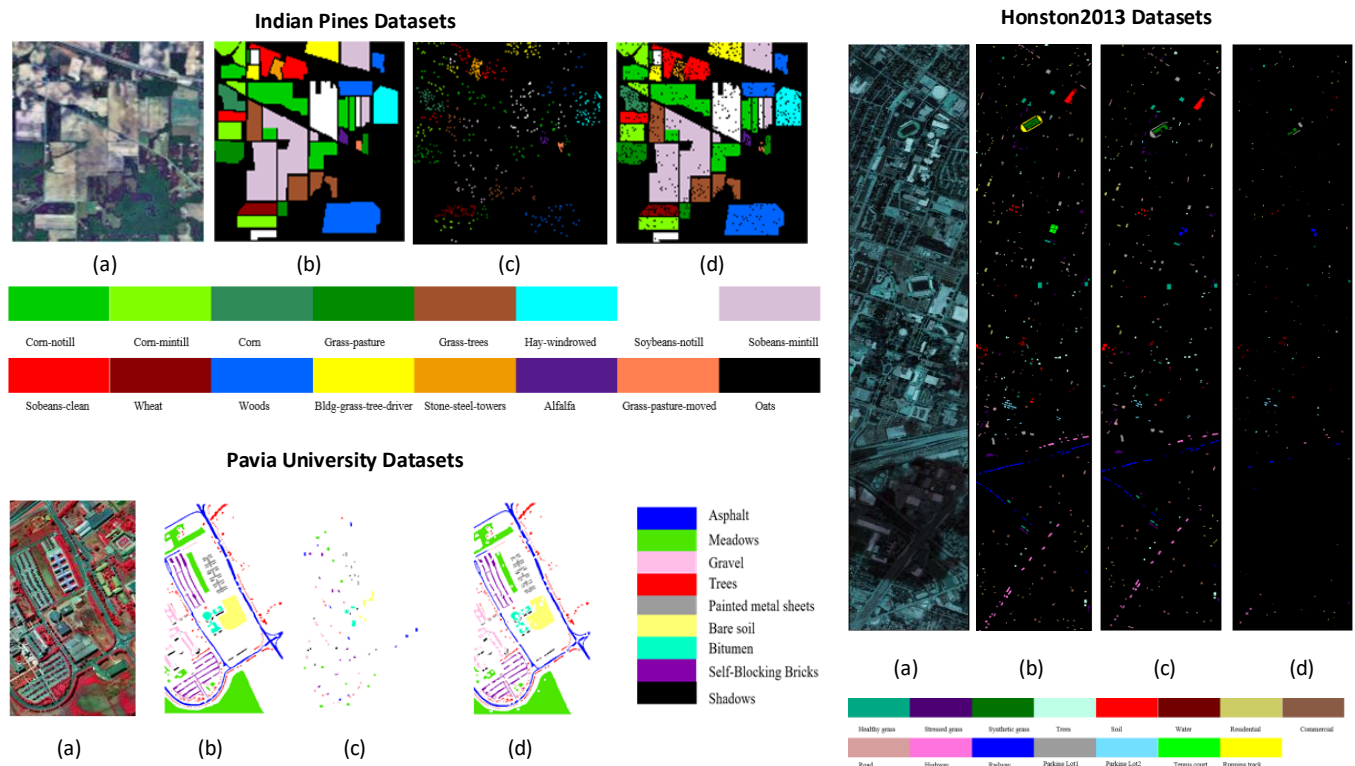

Fig. 3. Hyperspectral data sets. (a) False-color composite image, (b) Ground-truth map, (c) Training samples, (d) Test samples.

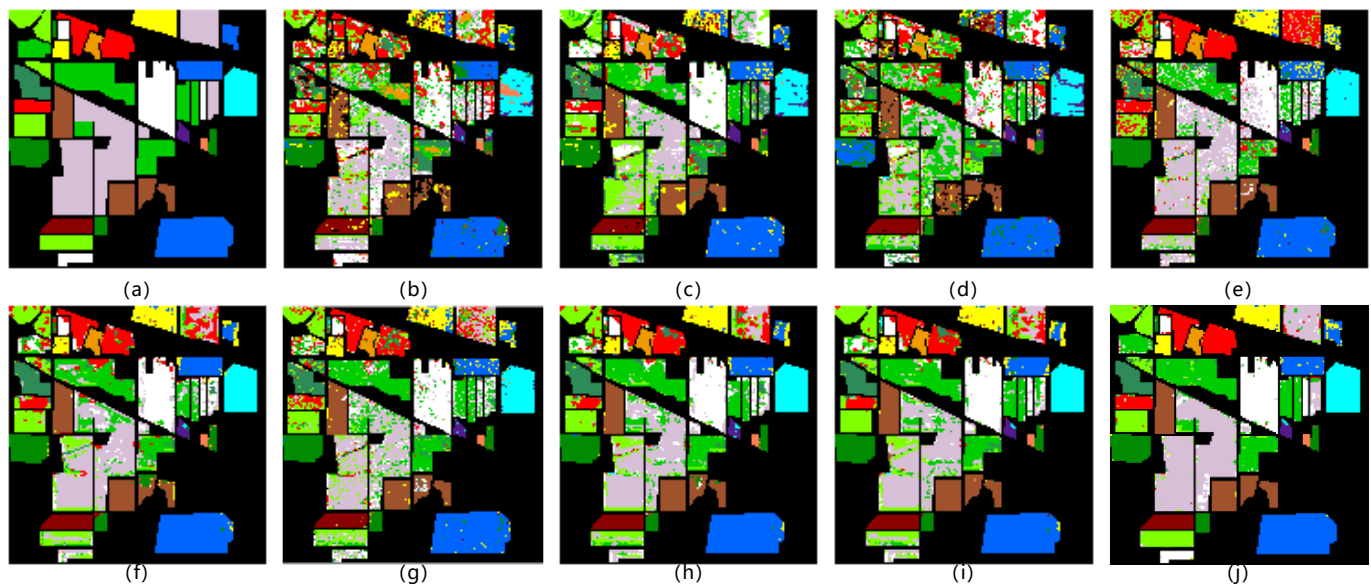

Fig. 4. Classification maps obtained by different methods on the Indian Pines data set. (a) Ground-truth map, (b) RF, (c) SVM, (d) 1D CNN, (e) 2D CNN, (f) 3D CNN, (g) GCN, (h) FuNet-C, (i) A-SPN, (j) MSHCNet.

### B. Experimental Settings

*1) Implementation Details:* As shown in Fig. 1, the MSHCNet architecture consists of four streams and a feature fusion part. For the G-stream and S-stream in MSHCNet, the number K of each KNN graph was empirically set as 10. We used $\ell$-norm regularization, set to 0.001, employing on weights to stabilize the network training and reduce overfitting. In the feature fusion part, the predicted MLP contained successive 1D Convs, each with 512, 128, and *P* channels, respectively. All 1D Convs operations were followed by batch normalization and LeakyReLU, except the last one MLP, which was followed by a tensor-reshape operation to output the probability matrix of each pixel.

Our MSHCNet was trained by minimizing the cross-entropy classification loss on NVIDIA GTX 1080 GPUs for 200 epochs using the Tensorflow platform. Analogous to [39], we utilized the minibatch size setting as 7. The initial learning rate was 1e-3, which was reduced by 0.5 decay at intervals of 50 epochs. The used codes and corresponding results in our experiments were available at https://github.com/NZWANG/MSHCNet.

*2) Competing Methods:* Our MSHCNet was compared with other eight state-of-the-art methods for HSI classification, including RF [15], SVM [13], 1D CNN [23], 2D CNN [26], 3D CNN [28], GCN [24], FuNet-C [39], and A-SPN [34]. All these competing methods were implemented by their original codes and trained on the same HSI datasets. The overall classification performance (averaged over all classes) was quantitatively evaluated by three widely used metrics, i.e., overall accuracy (OA), average accuracy (AA), and kappa coefficient ($\kappa$).

TABLE V
CLASSIFICATION ACCURACIES OBTAINED BY DIFFERENT METHODS ON THE INDIAN PINES
DATA SET. THE BEST ONE IS SHOWN IN BOLD.

| Evaluation metrics | | RF | SVM | 1D CNN | 2D CNN | 3D CNN | GCN | FuNet-C | A-SPN | MSHCNet |
|---|---|---|---|---|---|---|---|---|---|---|
| Per-class accuracy | 1 | 58.92 | 65.48 | 50.22 | 63.65 | 67.92 | 66.67 | 74.13 | 75.71 | **76.99** |
| | 2 | 57.65 | 71.31 | 42.22 | 73.59 | 72.93 | 72.77 | 77.93 | 74.25 | **79.37** |
| | 3 | 80.02 | 94.23 | 59.64 | 96.19 | 93.48 | 87.55 | 97.28 | 92.33 | **100** |
| | 4 | 84.88 | 92.21 | 89.49 | 93.06 | 92.62 | 93.04 | 94.85 | 93.09 | **94.88** |
| | 5 | 79.96 | 89.01 | 93.40 | 88.52 | 95.70 | 90.39 | 95.70 | 95.94 | **96.43** |
| | 6 | 94.61 | 93.21 | 97.04 | 98.63 | 99.09 | 97.63 | 99.77 | 98.83 | **100** |
| | 7 | 77.63 | 74.05 | 65.47 | 77.12 | 64.16 | 75.32 | 79.74 | 73.31 | **85.75** |
| | 8 | 58.61 | 53.68 | 64.97 | 62.15 | 67.99 | 50.71 | 63.19 | 68.37 | **74.34** |
| | 9 | 60.19 | 73.42 | **93.79** | 80.67 | 72.70 | 62.37 | 76.24 | 82.51 | 83.63 |
| | 10 | 96.78 | 97.59 | 99.38 | **100** | 99.38 | 97.63 | **100** | 99.14 | **100** |
| | 11 | 89.49 | 87.34 | 84.49 | 93.56 | 90.27 | 86.52 | 92.44 | **98.52** | 96.22 |
| | 12 | 52.57 | 71.63 | 85.15 | 89.69 | 90.46 | 67.69 | 90.91 | 89.41 | **93.67** |
| | 13 | 92.13 | 95.37 | 93.33 | **100** | 97.78 | 96.51 | **100** | 96.42 | **100** |
| | 14 | 31.17 | 81.65 | 84.62 | 79.48 | 74.33 | 70.95 | 76.92 | 79.28 | **87.18** |
| | 15 | 80.27 | 90.02 | **100** | **100** | **100** | 80.82 | **100** | 95.86 | **100** |
| | 16 | 46.24 | 97.59 | 70.00 | **100** | **100** | 98.42 | **100** | **100** | **100** |
| OA(%) | | 68.24 | 74.36 | 71.31 | 77.35 | 76.99 | 75.88 | 80.52 | 80.87 | **83.23** |
| AA(%) | | 71.32 | 82.98 | 79.58 | 87.27 | 86.17 | 80.94 | 88.69 | 88.31 | **91.77** |
| $\kappa \times 100$ | | 64.85 | 69.93 | 67.41 | 74.37 | 73.98 | 70.88 | 78.39 | 77.96 | **80.72** |

*C. Classification Results*

To show the effectiveness of our proposed MSHCNet, here we quantitatively and qualitatively evaluated the classification performance by comparing with the aforementioned methods.

*1) Results on the Indian Pines Hyperspectral Dataset*

The quantitative results obtained by different methods on the Indian Pines dataset were summarized in Table V, where the highest value in each row was highlighted in bold. Overall, RF and SVM obtained similar classification performance, where this might be explained by the relatively low robustness to a few noisy training samples. Being beneficial from the powerful learning ability of deep learning techniques, the CNN-based and GCN-based methods performed better than traditional classifiers (RF and SVM). However, we observed that the CNN-based ones including 1D CNN and 2D CNN achieved relatively low accuracy, which was due to that they can only conduct the convolution on a regular image grid, so the specific local spatial information cannot be captured. Unlike 1D CNN and 2D CNN, 3D CNN can extract the spatial-spectral information from HSIs more effectively, yielding higher classification accuracies. We had to point out, however, that the 3D CNN required additional network parameters to be estimated and tended to suffer from overfitting problems. By contrast, GCN-based methods were capable of adaptively aggregating the features on irregular non-Euclidean regions, so they could yield better performance than CNN-based and traditional methods. The A-SPN model, which combined spectral and spatial information in second-order statistics, ranked in the high place. This implied that the second-order features were quite useful to enhance the classification performance. Moreover, the FuNet-C (that combined the benefited of CNNs and GCNs) outperformed those single models, demonstrating its ability to fuse different spectral representations. Furthermore, we observed that the proposed MSHCNet achieved the top level performance among all the methods in terms of OA, AA, and Kappa coefficient.

Fig. 4 exhibited a visual comparison of the classification results generated by different methods on the Indian Pines dataset. Compared with the ground-truth map in Fig. 5(a), it can be seen that some pixels of 'Soybean-mintill' were misclassified into 'Corn-notill' in all the classification maps because these two land-cover types had similar spectral

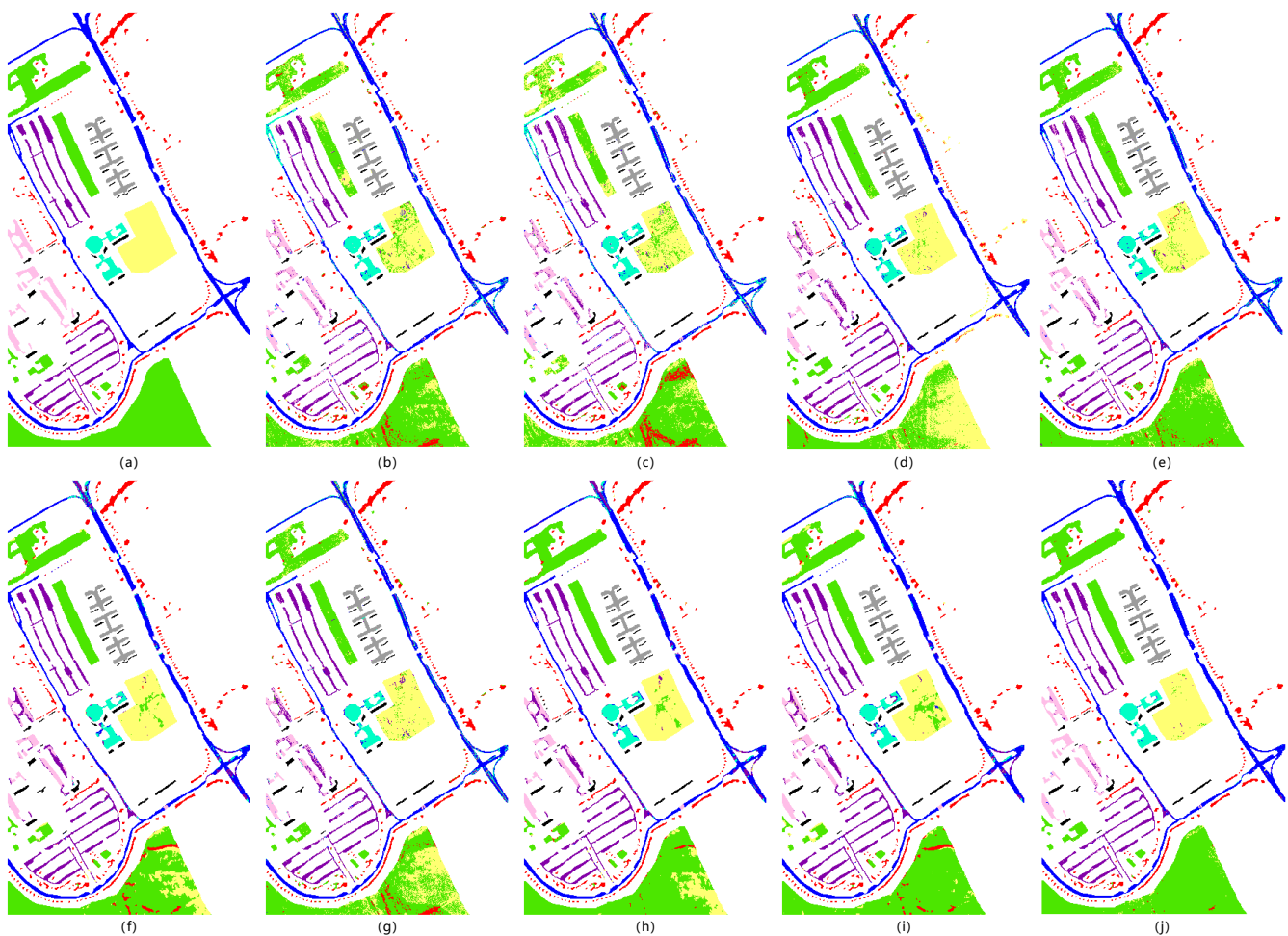

Fig. 5. Classification maps obtained by different methods on the Pavia University data set. (a) Ground-truth map, (b) RF, (c) SVM, (d) 1D CNN, (e) 2D CNN, (f) 3D CNN, (g) GCN, (h) FuNet-C, (i) A-SPN, (j) MSHCNet.

TABLE VI
CLASSIFICATION ACCURACIES OBTAINED BY DIFFERENT METHODS ON THE PAVIA UNIVERSITY DATA SET. THE BEST ONE IS SHOWN IN BOLD.

| Evaluation metrics | | RF | SVM | 1D CNN | 2D CNN | 3D CNN | GCN | FuNet-C | A-SPN | MSHCNet |
|---|---|---|---|---|---|---|---|---|---|---|
| Per-class accuracy | 1 | 79.34 | 81.32 | 89.89 | 79.98 | 81.66 | 74.96 | 95.76 | 96.47 | **96.69** |
| | 2 | 56.91 | 68.77 | 57.18 | 80.71 | 90.12 | 69.41 | 96.96 | 96.63 | **97.58** |
| | 3 | 52.37 | 70.18 | 72.16 | 69.99 | 74.87 | 61.79 | 85.44 | 84.34 | **88.82** |
| | 4 | 80.86 | 80.38 | 80.27 | 93.36 | 95.45 | 98.35 | 90.13 | 83.44 | **98.90** |
| | 5 | 81.91 | 81.41 | 99.24 | 99.64 | 99.19 | 99.73 | 99.24 | **100** | **100** |
| | 6 | 80.10 | 84.36 | 91.32 | 91.85 | 91.71 | 82.72 | 89.26 | 90.37 | **92.03** |
| | 7 | 86.36 | 83.12 | 89.07 | 80.47 | 74.06 | 86.38 | 88.12 | 89.81 | **93.34** |
| | 8 | 90.31 | 91.25 | 87.14 | **95.32** | 94.78 | 93.11 | 90.97 | 92.53 | 94.62 |
| | 9 | 94.47 | 94.96 | 98.71 | 95.57 | 96.72 | 94.97 | 94.80 | 98.52 | **100** |
| OA(%) | | 70.98 | 73.53 | 78.32 | 85.69 | 87.55 | 78.82 | 90.17 | 91.06 | **93.14** |
| AA(%) | | 77.41 | 81.75 | 84.99 | 87.43 | 88.72 | 84.61 | 92.29 | 92.45 | **95.78** |
| κ × 100 | | 63.27 | 67.88 | 69.42 | 80.78 | 83.94 | 70.34 | 88.27 | 90.19 | **91.43** |

TABLE VII
CLASSIFICATION ACCURACIES OBTAINED BY DIFFERENT METHODS ON THE HOUSTON2013 DATA SET. THE BEST ONE IS SHOWN IN BOLD.

| Evaluation metrics | | RF | SVM | 1D CNN | 2D CNN | 3D CNN | GCN | FuNet-C | A-SPN | MSHCNet |
|---|---|---|---|---|---|---|---|---|---|---|
| Per-class accuracy | 1 | 82.83 | 82.98 | 87.31 | 85.09 | 84.70 | 90.13 | 85.76 | **95.71** | 94.05 |
| | 2 | 97.24 | 98.42 | 98.25 | **99.96** | 99.38 | 99.13 | 99.41 | 97.08 | 97.84 |
| | 3 | 78.02 | 79.55 | 78.78 | 77.18 | 84.57 | 79.89 | 80.74 | 98.72 | **98.84** |
| | 4 | 96.75 | 98.51 | 93.03 | 97.76 | 98.00 | 96.70 | 98.51 | 97.12 | **98.57** |
| | 5 | 97.11 | 97.82 | 97.30 | 98.51 | 97.87 | 86.14 | 99.29 | 97.28 | **99.66** |
| | 6 | 97.17 | 90.92 | 95.05 | 92.27 | 92.99 | 73.30 | 95.08 | 98.32 | **98.34** |
| | 7 | 81.89 | 90.43 | 77.31 | 92.17 | 86.28 | 95.12 | 91.61 | 82.15 | **96.81** |
| | 8 | 39.74 | 40.47 | 51.40 | 79.44 | 76.31 | 71.67 | 74.88 | 79.44 | **80.42** |
| | 9 | 68.97 | 41.91 | 27.99 | 86.34 | 84.28 | 70.91 | 85.25 | 78.00 | **86.59** |
| | 10 | 56.76 | 62.67 | 90.88 | 43.76 | 74.27 | 72.16 | 79.30 | 89.18 | **89.21** |
| | 11 | 77.04 | 75.40 | 79.35 | 87.04 | 82.37 | 87.22 | 82.30 | 81.80 | **88.08** |
| | 12 | 50.23 | 59.99 | 76.57 | 66.26 | 77.76 | 73.42 | 78.83 | 79.74 | **80.92** |
| | 13 | 61.44 | 49.48 | 69.44 | 92.16 | 91.08 | 86.39 | 89.10 | 94.06 | **95.09** |
| | 14 | 98.83 | 98.81 | 99.15 | 98.71 | 98.67 | 99.75 | 98.25 | 99.21 | **99.82** |
| | 15 | 96.97 | 97.42 | 98.05 | 97.75 | 94.62 | 99.35 | 96.66 | 99.13 | **99.43** |
| OA(%) | | 80.04 | 77.33 | 82.15 | 83.56 | 85.09 | 84.02 | 87.51 | 88.71 | **89.77** |
| AA(%) | | 78.73 | 77.65 | 81.24 | 86.29 | 88.21 | 85.41 | 89.00 | 88.80 | **93.57** |
| κ × 100 | | 77.93 | 76.30 | 80.42 | 82.31 | 84.15 | 82.61 | 85.83 | 87.83 | **89.92** |

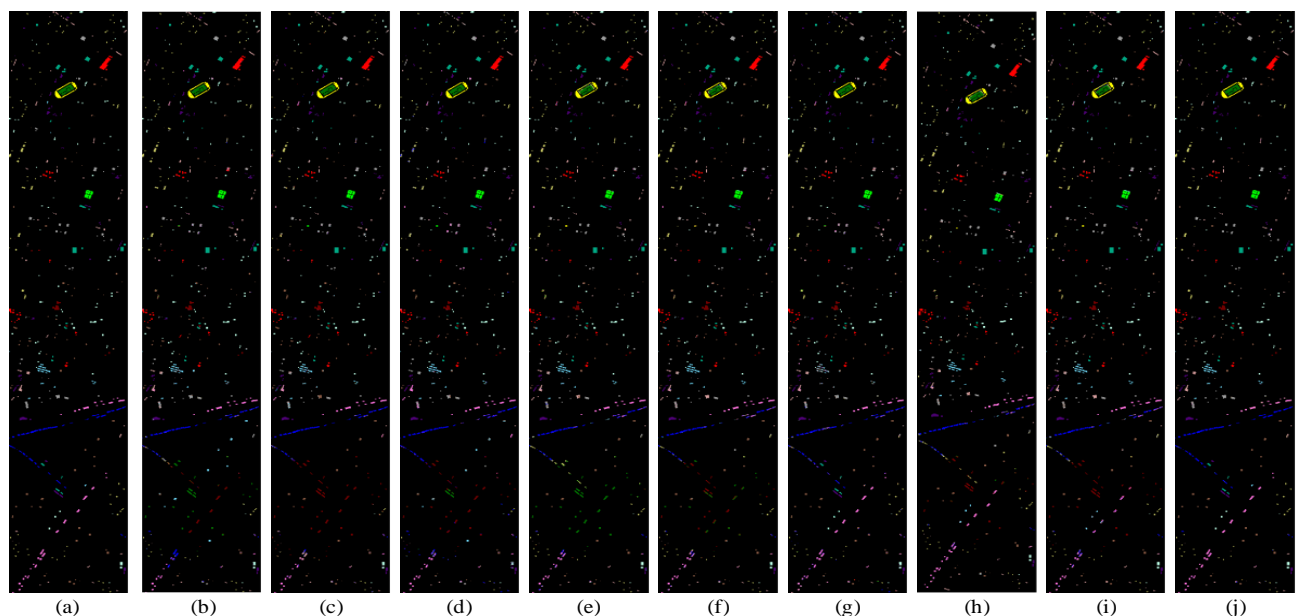

(a)　(b)　(c)　(d)　(e)　(f)　(g)　(h)　(i)　(j)

Fig. 6. Classification maps obtained by different methods on the Houston2013 data set. (a) Ground-truth map, (b) RF, (c) SVM, (d) 1D CNN, (e) 2D CNN, (f) 3D CNN, (g) GCN, (h) FuNet-C, (i) A-SPN, (j) MSHCNet.

signatures. Meanwhile, due to the lack of spatial context, the classification map obtained by GCN suffered from pepper-noise-like mistakes within certain regions. Comparatively, the results of the proposed MSHCNet yielded smoother visual effect and showed fewer misclassifications than other compared methods did.

*2) Results on the Pavia University Hyperspectral Dataset*

Table VI presented the quantitative comparison results of different methods on the Pavia University dataset. Similar to the results on the Indian Pines dataset, the results in Table VI indicated that the proposed MSHCNet ranked in the first place and outperformed the compared methods by a substantial margin, which again validated the strength of our proposed MSHCNet. Besides, it was also notable that FuNet-C performed better than A-SPN, which was different from the results on the Indian Pines dataset. This was mainly because that FuNet-C and our MSHCNet fused spectral-spatial information with diverse feature extractors, which could well adapt to the hyperspectral images containing many boundary regions. Since the objects belonging to the same class in the Pavia University dataset were often distributed in widely scattered and small regions, FuNet-C and MSHCNet were able to achieve better performance than A-SPN, GCN and other baseline methods. Furthermore, observed by Fig. 5, stronger spatial correlation and fewer misclassifications could be observed in the classification map generated by the proposed MSHCNet when compared with FuNet-C and other competitors.

*3) Results on the Houston2013 University Hyperspectral Dataset*

Table VII presented the comparison results of different methods on the Houston2013 University dataset. It was apparent that the performance of all methods was better than that on the Indian Pines and the Pavia University dataset. This could be due to that the Houston2013 University dataset had higher spatial resolution and contained less noise than the Indian Pines and the Pavia University dataset, and thus was more suitable for classification. As it could be noticed, A-SPN model achieved the highest OA and $\kappa$ among all the competing methods. However, slight gaps could still be observed between and our MSHCNet in terms of OA, AA and $\kappa$. For the proposed MSHCNet, it was also worth noting that all the classes had the higher accuracies and lower misclassifications, which further demonstrated the advantages of our proposed MSHCNet. Fig. 6 visualized the classification results of nine different methods. We could see that our MSHCNet was able to produce quite precise classification results on these small and difficult regions.

## IV. Discussion

### A. Effects of Different Neighbor Nodes in MSHCNet

The neighbor nodes scale $K$ in G-stream and S-stream could potentially affect the classification results of MSHCNet. To analyze its influence on our method, in this section, we respectively set the neighbor nodes scale $K$ to 5, 10, 15, 20, 25, and 30, and test their classification accuracies on each data set. The results of OA indices were shown in Fig. 7.

The scale of neighbor nodes indeed affected the performance of the constructed graph. The larger number of neighbor nodes ($K$) led to more nodes that preserved larger objects and suppresses more noise. As it could be seen from Fig. 7, the accuracy of the MSHCNet on the Indian Pines dataset increased first and then decreased with the increase of $K$, which indicated that there were more small objects existing in this HSI dataset. In contrast, the accuracy of MSHCNet on the Pavia University and Houston2013 dataset only yielded a slight performance fluctuation, indicating that the HSI may consist of more large and small objects at the same time. To prevent the MSHCNet from obtaining oversmoothed classification maps, we empirically fixed it to 10 in all experiments.

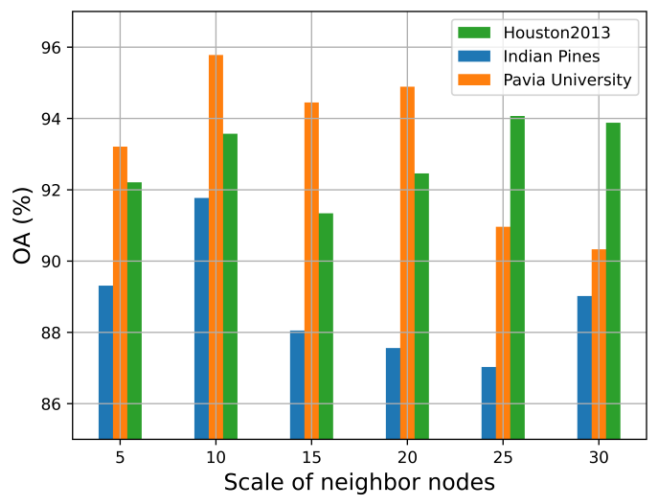

Fig. 7. Classification overall accuracy of MSHCNet with different neighbor nodes on Indian Pines, Pavia University and Houston2013 datasets, respectively.

TABLE VIII
CLASSIFICATION RESULTS WITHOUT VARIOUS COMPONENTS ON EACH DATASET.

| Datasets | | Without C-stream | Without G-stream | Without N-stream | Without S-stream | MSHCNet |
|---|---|---|---|---|---|---|
| Indian Pines | OA (%) | 75.37 | 80.20 | 81.49 | 80.77 | **83.23** |
| | AA (%) | 79.08 | 87.33 | 89.61 | 88.34 | **91.77** |
| | $\kappa \times 100$ | 70.64 | 75.46 | 78.63 | 77.28 | **80.72** |
| Pavia University | OA (%) | 76.41 | 89.06 | 91.02 | 91.04 | **93.14** |
| | AA (%) | 85.73 | 90.70 | 92.28 | 90.88 | **95.78** |
| | $\kappa \times 100$ | 72.31 | 87.05 | 89.58 | 88.52 | **91.43** |
| Houston2013 | OA (%) | 70.98 | 88.17 | 85.23 | 80.28 | **89.77** |
| | AA (%) | 71.46 | 91.33 | 87.82 | 84.33 | **93.57** |
| | $\kappa \times 100$ | 69.17 | 88.31 | 85.68 | 82.78 | **89.92** |

*B. Effects of Different Stream Components of MSHCNet*

In order to explore each stream's affecting performance of MSHCNet, we designed four variations: MSHCNet without C-stream, MSHCNet without G-stream, MSHCNet without N-stream, and MSHCNet without S-stream. Note that we only analyzed the effects of each component so that the usage of other parameters was referred to previous work. Table VIII illustrated the OA, AA and $\kappa$ of MSHCNet based on three datasets.

As listed in Table VIII, without C-stream, G-stream, N-stream, or S-stream, MSHCNet, precisely, boiled down to a very unstable level, and the OA, AA and $\kappa$ relied heavily on the CNN layer, i.e., the OA, AA and $\kappa$ are decreased by at least 8% without C-stream, respectively. In addition, without G-stream, N-stream or S-stream, the OA, AA and $\kappa$ of MSHCNet were all affected to a greater or lesser extent. Moreover, Houston2013 dataset was more sensitive to S-stream, whereas Pavia University was more stable. For instance, MSHCNet obtained an OA of 80.28% for the Houton2013 dataset decreased by 9.49% without S-stream, whereas the OA obtained by the Pavia University was decreased by only 2.1%. Accordingly, the combination of CNN and GCN with the mixed statistics could be more superior over other variants, and it was more stable on each dataset.

## V. CONCLUSION

A multi-stream hybridized convolutional network with mixed statistics in Euclidean/Non-Euclidean spaces, called MSHCNet, has been proposed in this article for the automatical HSI classification. Specifically, our MSHCNet first adopted four parallel feature representation streams, which contained G-stream, utilizing the irregular correlation between adjacent land covers in non-Euclidean space; C-stream, adopting convolution operator to learn regular spatial-spectral features in Euclidean space; N-stream, combining first and second order features to learn representative and discriminative regular spatial-spectral features in Euclidean space; S-stream, using GSOP to capture boundary correlations and obtain graph representations from all nodes in non-Euclidean space. Specifically, being different from prior works that depended only on the first-order spectral signatures, It can be noted that the proposed MSHCNet critically employed GSOP operator, which extracted second-order feature correlation making boundary information more powerful, for HSI classification. Therefore, this feature representation network can faithfully find discriminative and representative spatial-spectral features of local regions, and help to find accurate representations. Furthermore, the feature representations produced by four different streams were proposed to fuse the complementary multi-view information by MLP for the end-to-end pixel-wise prediction. An extensive comparison has been performed between our MSHCNet and other eight state-of-the-art methods on three public HSI datasets, and the corresponding results demonstrated the superiority of our proposed method, especially for small and local areas.

Despite the encouraging experimental performance based on our proposed approach, we plan to extend our current MSHCNet for follow-up research from the following two aspects: 1) explore higher-order information in design of the novel effective CNN and GCN architectures; 2) investigate the attention network with new architecture for addressing the HSI class-imbalance problem.

ACKNOWLEDGMENT

The authors would like to thank Prof. D. Landgrebe for making the Airborne Visible/Infrared Imaging Spectrometer Indian Pines hyperspectral data set available to the community, Prof. P. Gamba for providing the Reflective Optics Spectrographic Imaging System data over Pavia, Italy, and the IEEE Geoscience and Remote Sensing Society (GRSS) Data Fusion Technical Committee for providing the University of Houston data sets.